%% file: main.tex
\documentclass[10pt,twocolumn,letterpaper]{article}
\pdfoutput=1
\usepackage[pagenumbers]{cvpr} 

\usepackage{multirow}

\usepackage{algorithm}
\usepackage{algpseudocode}

\algtext*{EndIf} 
\algtext*{EndFor} 
\algtext*{EndWhile} 

\usepackage{makecell}
\usepackage{arydshln}
\usepackage[most]{tcolorbox} 
\usepackage{array}
\usepackage{tabularx} 
\usepackage{caption} 
\input{preamble}

\usepackage{lineno}     

\definecolor{cvprblue}{rgb}{0.21,0.49,0.74}
\usepackage[pagebackref,breaklinks,colorlinks,allcolors=cvprblue]{hyperref}

\def\paperID{4275} 
\def\confName{CVPR}
\def\confYear{2025}

\title{GUI Testing Arena: A Unified Benchmark \\for Advancing Autonomous GUI Testing Agent}

\author{%
    Kangjia Zhao$^{1}$ \quad
    JiaHui Song$^{1}$ \quad
    Leigang Sha$^{1}$ \quad
    HaoZhan Shen$^{1}$ \\
    Chen Zhi$^{1}$\thanks{Corresponding author} \quad
    Tiancheng Zhao$^{2,3}$ \quad
    Xiubo Liang$^{1}$ \quad
    Jianwei Yin$^{1}$\\[3pt]
    $^1$ College of Computer Science and Technology, Zhejiang University\\
    $^2$ Om AI Research \quad $^3$ Binjiang Institute of Zhejiang University \\
    {\tt\small \{konkaz, songjah, shaleigang, hz$\_$shen, zjuzhichen, xiubo, zjuyjw\}@zju.edu.cn}\\
    {\tt\small tianchez@zju-bj.com}
    \vspace{-3mm}
}
\begin{document}
\maketitle
\input{sec/0_abstract}    
\input{sec/1_intro}
\input{sec/2_Workflow}
\input{sec/3_Benchmark}

\input{sec/4_Correlation}

\input{sec/5_Experiment}
\input{sec/6_Related_Work}
\input{sec/7_Conclusion}
{
    \small
    \bibliographystyle{ieeenat_fullname}
    \bibliography{main}
}


\begin{appendix}
\input{sec/Appendix}
\end{appendix}

\end{document}

%% file: preamble.tex
\usepackage{anyfontsize}
\usepackage{listings}
\definecolor{codeblue}{rgb}{0.25,0.5,0.5}
\definecolor{codekw}{rgb}{0.85, 0.18, 0.50}
\definecolor{Highlight}{HTML}{39b54a}
\definecolor{grayL}{RGB}{200, 200, 200}

\newcolumntype{x}[1]{>{\centering\arraybackslash}p{#1pt}}
\newcolumntype{y}[1]{>{\raggedright\arraybackslash}p{#1pt}}
\newcolumntype{z}[1]{>{\raggedleft\arraybackslash}p{#1pt}}

\newcommand{\tablestyle}[2]{\setlength{\tabcolsep}{#1}\renewcommand{\arraystretch}{#2}\centering\footnotesize}

\newlength\savewidth
\newcommand\shline{\noalign{\global\savewidth\arrayrulewidth
  \global\arrayrulewidth 1pt}\hline\noalign{\global\arrayrulewidth\savewidth}}

\newcommand{\tagfont}[1]{{\fontsize{7pt}{1em}\selectfont{\textcolor{grayL}{#1}}}}
\newcommand{\ressup}[2]{\tablestyle{1pt}{1} \begin{tabular}{z{18}y{22}} {#1} & {} \end{tabular}}
\newcommand{\reshl}[3]{\tablestyle{1pt}{1} \begin{tabular}{z{18}y{22}} {#1} & \fontsize{7.5pt}{1em}\selectfont{\textcolor{Highlight}{(${#2}$\textbf{#3})}} \end{tabular}}

%% file: sec/0_abstract.tex
\begin{abstract}
   Nowadays, research on GUI agents is a hot topic in the AI community. However, current research focuses on GUI task automation, limiting the scope of applications in various GUI scenarios. In this paper, we propose a formalized and comprehensive environment to evaluate the entire process of automated GUI Testing (GTArena), offering a fair, standardized environment for consistent operation of diverse multimodal large language models.  We divide the testing process into three key subtasks: test intention generation, test task execution, and GUI defect detection, and construct a benchmark dataset based on these to conduct a comprehensive evaluation. It evaluates the performance of different models using three data types: real mobile applications, mobile applications with artificially injected defects, and synthetic data, thoroughly assessing their capabilities in this relevant task. Additionally, we propose a method that helps researchers explore the correlation between the performance of multimodal language large models in specific scenarios and their general capabilities in standard benchmark tests. Experimental results indicate that even the most advanced models struggle to perform well across all sub-tasks of automated GUI Testing, highlighting a significant gap between the current capabilities of Autonomous GUI Testing and its practical, real-world applicability. This gap provides guidance for the future direction of GUI Agent development. Our code is available at \href{https://github.com/ZJU-ACES-ISE/ChatUITest}{https://github.com/ZJU-ACES-ISE/ChatUITest}.
\end{abstract}


%% file: sec/1_intro.tex
\section{Introduction}
\label{sec:intro}

\begin{figure*}[!t]
    \centering
    \includegraphics[width=15cm]{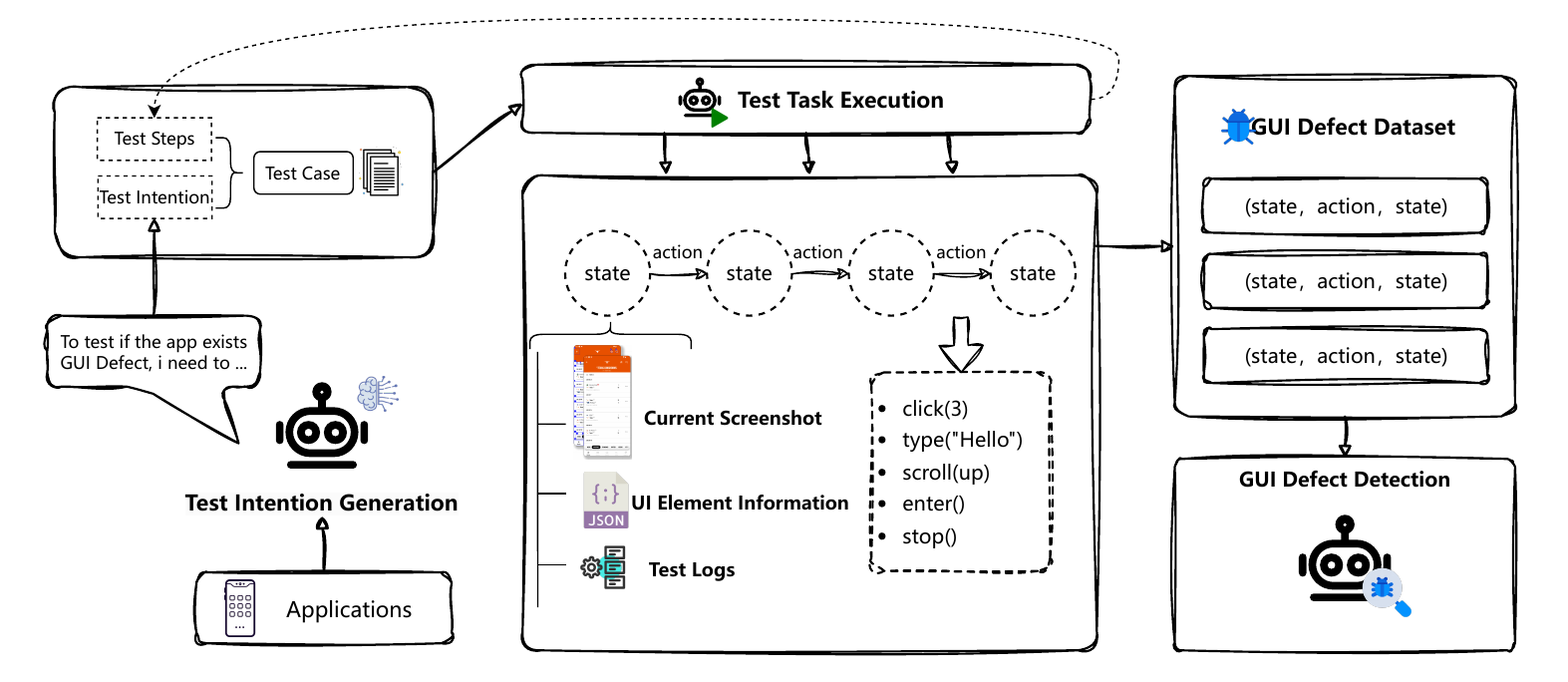}
    \caption{The Workflow for Autonomous GUI Testing (GTArena). GUI Testing requires the model to perform specific tasks, all of which are evaluated within this workflow. We provide a standardized and reproducible testing framework, enabling fair comparison of different multimodal large language models.}
    \label{fig:workflow}
\end{figure*}

\emph{``Imagine a software test engineer casually leaning back in their chair, saying, `Alright, test this app for me' and an agent springs into action—generating test cases, executing tasks, sniffing out bugs—and delivers a complete report, all without breaking a sweat.''}

\vspace{10pt}

This envisioned scenario, where agents augment human efforts in software testing, is becoming increasingly plausible as \emph{large language models} (LLMs) \cite{chowdhery2023palm, touvron2023llama, ouyang2022training} and \emph{vision LLMs} (VLLMs) \cite{achiam2023gpt, alayrac2022flamingo, bai2023qwenvl,liu2024visual} emerge as potent tools for automating complex processes. The integration of traditional agents with the cognitive capabilities of LLMs or VLLMs represents a cutting-edge direction in contemporary research, with numerous studies \cite{ma2024comprehensive, hong2024cogagent, cheng2024seeclick, putta2024agent} focusing on enhancing navigation frameworks for web and app interfaces. Current methodologies predominantly revolve around singular tasks like \texttt{\small <assist me with a purchase>} or \texttt{\small <log in and post a tweet>}, which are executed by these agents.

However, these applications often suffer from a narrow focus, with task complexity increased only through ambiguous instructions or added steps, approaches that do not fundamentally enhance the agent’s capabilities. This limitation highlights the potential application of these technologies in automated GUI Testing, a field of considerable practical importance and complexity, which presents comprehensive challenges to the capabilities of current agents.

Automated GUI Testing using LLMs \cite{liu2023chatting, droidbot-gpt, liu2024make} has gained substantial traction in recent research. Recently, the focus has shifted towards leveraging MLLMs in place of previous agents, enabling these models to ``see'' and interact with GUI elements visually. This approach has garnered attention in both the AI research community \cite{you2025ferret, you2024ferret} and the software engineering (SE) community \cite{liu2024vision, hu2024auitestagent}, leading to various exploratory studies and small-scale validations in industry. \textbf{As envisioned at the beginning of the paper, the ultimate goal of this research direction is to achieve end-to-end automation in GUI Testing.}

However, existing agents for automated GUI Testing tend to rely on complex framework designs, introducing various components to accomplish this task. Additionally, these frameworks lack standardized evaluation metrics, with variations in metric design and effectiveness measurement, and the test datasets are often small with limited availability for open access. This lack of standardization and transparency hinders further progress in GUI Testing research.

Moreover, due to the complexity of agent frameworks and the variability in responses of LLMs, reproducing results becomes challenging for others who want to evaluate the agent. Most current agent frameworks depend on GPT \cite{achiam2023gpt}, which are not only costly but also lack options for local deployment. They remain unsuitable for applications requiring data privacy, presenting an additional barrier to their widespread use in GUI Testing.

To address these gaps, we formalize the GUI automated testing problem by redefining the entire testing process with standard definitions and building a framework (GTArena) for fair GUI Testing evaluation. We introduce a novel data structure for representing GUI defects, enabling the construction of large-scale GUI defect datasets for future research. We formalize the entire GUI workflow, making agent evaluation standardized and reproducible. Building on this foundation, we have developed a unified benchmark for evaluating visual-based agents across various \emph{multimodal large language models} (MLLMs). This benchmark not only assesses the end-to-end task performance but also evaluates each component of the process individually, demonstrating that current agents still fall short in completing these tasks. Through this approach, we can analyze the specific performance gaps between GPT-based agents and other multimodal large models, providing insights that may guide future improvements.Furthermore, while benchmarks for multimodal large models typically assess specific capabilities, when applying these models in specific domains, only the end metrics are considered. Therefore, we propose a new method: evaluating models fine-tuned on subsets of datasets on general capability benchmarks and comparing these results to the original models. This analysis aids in identifying the necessary enhancements for the models to excel in specific tasks, facilitating further refinement of their capabilities.

The contributions of this paper are as follows:
\begin{itemize}
    \item We establish a formalized end-to-end framework as a fair environment for Autonomous GUI Testing (GTArena) and introduce a novel data structure for GUI defects, which together redefine the complete testing process and enable the construction of large-scale GUI defect datasets. This dual development not only decouples the agent but also facilitates rigorous and reproducible evaluations of GUI Testing methodologies.
    \item We develop a comprehensive and standardized benchmark for evaluating agents across multiple components within the automated GUI Testing framework, providing detailed insights into the agents' performance relative to human testers.
    \item We propose a methodology for assessing the specific capabilities that models need to excel in Autonomous GUI Testing, enabling targeted enhancements of these models through broader and more diverse training datasets.
\end{itemize}

%% file: sec/2_Workflow.tex
\section{Autonomous GUI Testing Agent Workflow}
\label{agent_workflow}
While the idea of casually handling over an app to an agent with a simple \texttt{\small <Test this app for me>} sounds appealing, the reality of fully automated GUI Testing is far more intricate. In order to effectively approach automated GUI Testing, it is crucial to deconstruct the testing workflow much like a skilled GUI Testing engineer would. The initial step involves defining testing objectives—primarily identifying GUI defects that pose the highest risk to user experience. This process begins with the predefinition of core tasks that reflect the most likely user interactions within an app. By executing these tasks on the app interface and closely monitoring for GUI defects, a comprehensive assessment can be achieved.
When leveraging agents to simulate this structured testing process, the workflow of automated GUI Testing can be divided into three main phases: \textbf{Test Case Generation}, where potential user interactions and testing scenarios are designed; \textbf{Test Task Execution}, in which the agent performs these tasks across the GUI; and \textbf{GUI Defect Detection}, a critical phase to identify any interface issues that may impair usability or functionality. The specific workflow process is presented in Figure \ref{fig:workflow}.

\subsection{Preliminary}

Current research on Automated GUI Testing predominantly concentrates on resolving specific issues within the domain. However, a rigorous definition and structured framework have been largely absent. To address this gap, we have formalized the process of this task through a Visual-based Agent, offering a novel perspective that redefines the task. Algorithm \ref{alg:code} provides the pseudocode implementation of the architecture.

\subsubsection{Partially Observable Markov Decision Process}

The cornerstone of our framework is the \emph{partially observable markov decision process} (POMDP), which serves as the foundational model for describing the decision-making process of the Visual-based Agent in GUI Testing scenarios. A POMDP is defined by a tuple $(S,O,A,T,R)$, where $S$ denotes the state space of the application's GUI, $O$ represents the observation space, $A$ is the set of possible actions, $T: S \times A \rightarrow S$ is the transition function mapping actions in states to probability distributions over states, and $R: S \times A \times S \rightarrow R$ is the reward function. In the context of Autonomous GUI Testing, $O$ denotes a partial observation of the app’s current state. Due to inherent limitations, a GUI agent cannot fully capture all state information, particularly for closed-source applications where key elements, such as the Accessibility Tree, are inaccessible. This partial observability impacts the likelihood of detecting issues during GUI Defect Detection, as the agent relies on limited feedback to infer potential defects. Detection likelihood, therefore, depends on whether a defect is observable following a specific action. To address this challenge, the reward function $R$ is designed to assign positive rewards for successfully detecting GUI defects, incentivizing the agent to prioritize actions that uncover critical issues affecting user experience. This approach promotes the identification and resolution of impactful defects, despite the agent’s incomplete view of the app’s underlying state.

\subsubsection{GUI Defect Data Model}

To systematize the identification and classification of GUI defects, we introduce a novel data structure termed the Transition Tuple $(state_b, action,state_a)$. This tuple effectively captures the GUI state before ($state_b$) and after ($state_a$) an action is executed, with the action itself represented in the middle. A sequence of such tuples forms a complete path of state transitions within the application. We define a specialized action $\varnothing$, distinct from standard agent operations, to signify performing no operation on the app.

We have designed two classes to support our data model: the State class, which encapsulates methods init for initializing a state and act for performing an action, and the Transition Tuple class, with methods init to create instances and check to evaluate transitions for defects. This classification aids in formalized defect detection, allowing for delayed, yet comprehensive defect analysis without the need for real-time feedback.

\begin{algorithm}[!t]
\caption{Pseudocode of Workflow}
\label{alg:code}
\begin{lstlisting}[language=python]
class State:
    def act(self, action):
        # Returns a new state based on the action
        return new_state

class TransitionTuple:
    def __init__(self, state_before, action, state_after):
        self.state_before = state_before
        self.action = action
        self.state_after = state_after

    def check_defect(self):
        # Checks and returns whether there's a defect
        return defect_found

# Process for simulating the automated GUI Testing
def simulate_gui_testing(initial_state, actions):
    for action in actions:
        next_state = initial_state.act(action)
        transition = TransitionTuple(initial_state, action, next_state)

        if transition.check_defect():
            log_defect(transition)

        initial_state = next_state  # Update the current state

def log_defect(transition):
    # Log the defective transition
    print("Defect detected in transition:", transition)

# Usage example
actions = ['click', 'scroll', 'type']
initial_state = State()
simulate_gui_testing(initial_state, actions)

\end{lstlisting}
\end{algorithm}
\subsection{Test Case Generation}\label{test_case_generation}

Test Case Generation can be viewed as the process of creating a structured chain of states and actions aligned with the testing intentions. Given a mobile application, the agent begins by gathering a brief overview of the app and creating test intentions, defining what needs to be tested. Once the test intention is defined, the agent can execute the testing tasks outlined in Section \ref{task_execution}, generating a comprehensive test case that includes both the test intention and corresponding test steps. This structured approach enables the agent to conduct targeted and informed testing on the same app in future scenarios, such as version updates or feature enhancements. By leveraging these predefined test cases, the agent can focus on high-priority areas and adapt its testing to ensure new changes align with expected functionality, making the testing process both efficient and scalable.

\subsection{Test Task Execution}\label{task_execution}

In executing the detailed test case, the multimodal agent performs several key actions to interact with the mobile application. The primary actions include \textit{Click}, \textit{Scroll}, and \textit{Type}, along with the additional actions \textit{Stop} and \textit{Enter}. The execution starts with the agent activating the initial state of the application. As the agent interacts with the interface, it records every action, capturing screenshots and logging each step. This thorough recording process ensures that the impact of each action is documented, allowing for a detailed assessment of the app's response to user interactions. As the agent progresses through the test case, it navigates various screens and functionalities of the application. The execution phase continues until one of two outcomes occurs: task completion or a problem, using \textit{Stop} to represent. 

This structured approach to executing test cases, combined with the agent’s ability to record and react to the application’s state. This systematic execution process is essential for GUI automation testing, ensuring that multimodal large models can effectively mimic human testers in testing mobile applications.

\subsection{GUI Defect Detection}

In GUI automation testing, defect detection is essential for ensuring application quality and usability. GUI defects can be broadly classified into two main categories: \textbf{Display Defects} (DD) and \textbf{Interaction Defects} (ID). \cite{lelli2015classifying, xiong2023empirical, liu2024empirical}  
Display Defects focus on the visual presentation of the UI and include:

\begin{itemize} \item \textbf{Data Display}, such as Content Error and Data Type or Format Error. \item \textbf{Layout}, including UI Element Missing, UI Element Overlapping, Alignment Issues, and Uneven Spacing. \item \textbf{Style}, such as Inconsistent Color, Inconsistent Element Size, and Abnormal UI Element State. \end{itemize}

\noindent Interaction Defects pertain to user interactions with the UI and include:

\begin{itemize} \item \textbf{Operation}, like Operation No Response and Virtual Keyboard Related Issues. \item \textbf{Task}, including Navigation Logic Error and Unexpected Task Result. \end{itemize}

More details are provided in Appendix. By referencing these defect types, the agent can effectively detect and categorize any GUI defects present in each Transition Tuple and return the defect results. Through this automated process, the agent can verify that display and interaction elements perform optimally across various scenarios and user actions.

\begin{figure}[t!]
    \centering
    \includegraphics[width=8.5cm]{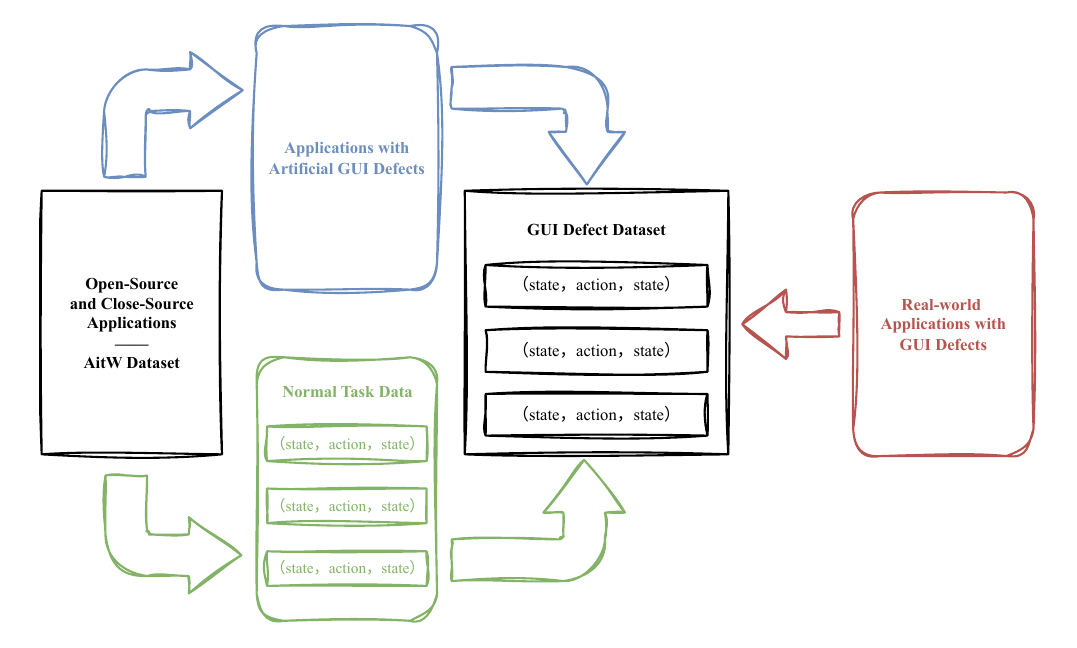}
    \caption{\textbf{Source and Methodology for Benchmark Data Construction.} The left side of the figure illustrates our primary data sources, which include intentionally injected defects within apps and synthetic defect data generated by post-processing action sequence data obtained from app executions. The right side of the figure shows supplemental data sources, specifically real-world applications with GUI defects.}
    \label{fig:data_source}
\end{figure}

\begin{table}
  \centering
  \begin{tabular}{@{}lcc@{}}
    \toprule
    Data of Applications & GUI Display & GUI Interaction \\  
    \midrule
    Real-World & \multicolumn{2}{c}{53} \\  
    \hline
    Artificial Inject & 79 & 26 \\  
    \hline
    AitW with Defects & 6421 & 1871 \\  
    Close-Source & 1148 & 399 \\  
    Open-Source & 590 & 257 \\  
    \bottomrule
  \end{tabular}
  \caption{Distribution of our dataset.}
  \label{tab:data_distribution}
\end{table}

\begin{figure*}[ht!]
    \centering
    \includegraphics[width=18cm]{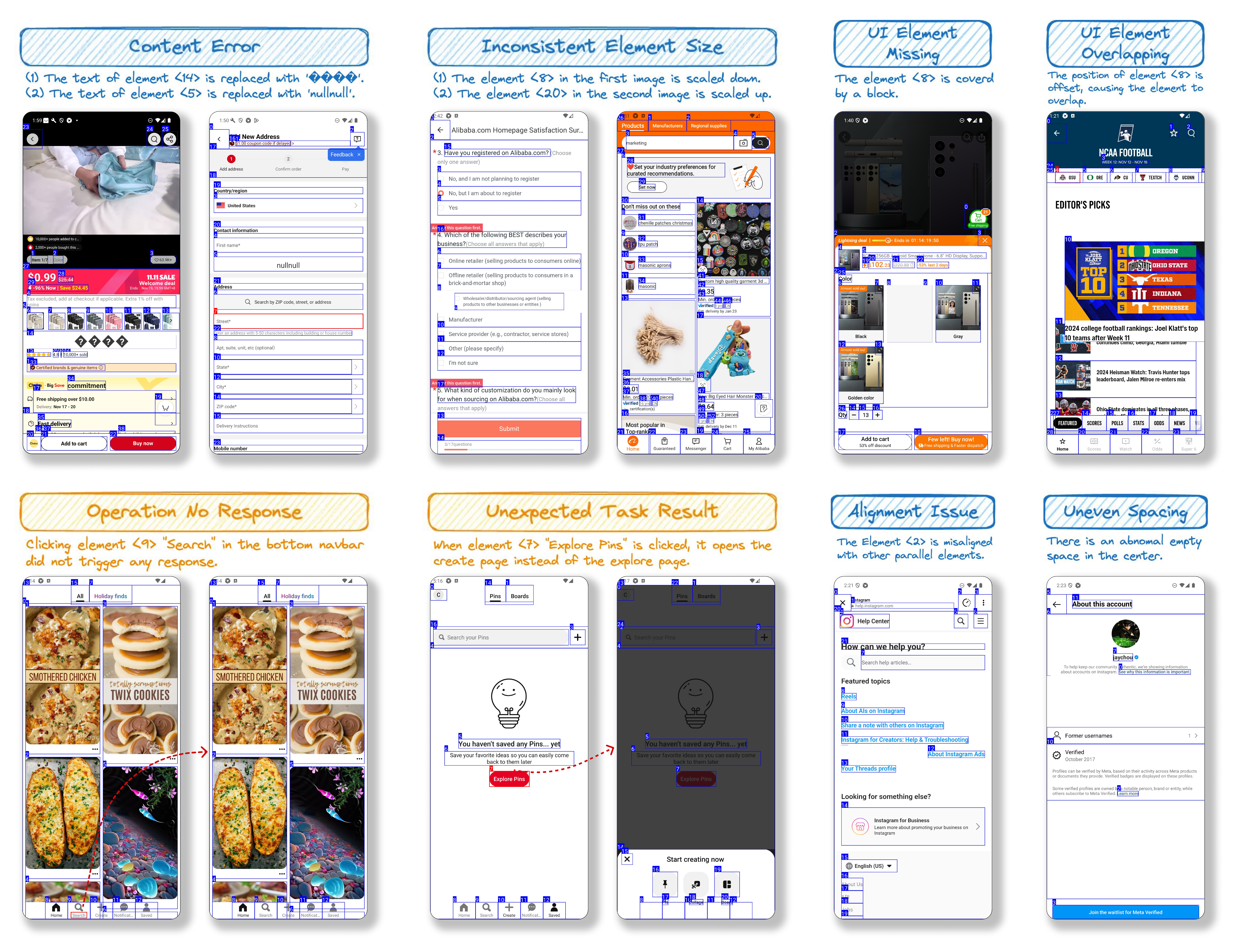}
    \caption{\textbf{Examples of Constructed Synthetic GUI Defects.} We present examples of various constructed GUI defects, demonstrating the feasibility of synthesizing defects through post-processing. This approach highlights a method for building large-scale GUI defect datasets, including both display and interaction defects.}
    \label{fig:synthesis_data}
\end{figure*}

%% file: sec/3_Benchmark.tex
\section{How To Benchmark Autonomous GUI Testing Agent}

Building a unified benchmark for the entire automated GUI Testing framework requires addressing challenges that arise from the segmented focus of prior research. Existing benchmark often isolate specific components, such as task  execution\cite{zhou2023webarena, bonatti2024windows} or defect detection\cite{alegroth2018continuous, su2021benchmarking}. Hence, we establish an end-to-end evaluation system, ensuring seamless integration across all phases of the testing process. 

Given the complexities involved in collecting data from real-world applications, where the defects are inherently unpredictable, our benchmark consists of three carefully curated data categories, including real-world mobile applications, applications with injected defects and synthetic defect datasets. Figures \ref{fig:data_source} illustrate the composition and of our benchmark, highlighting the balanced representation across each category. The detailed data distribution is shown in Table \ref{tab:data_distribution}.

\subsection{Real-world Applications with GUI Defects}
Real-world applications serve as a crucial component of our benchmark by providing insights into naturally occurring GUI defects. These defects are identified by mining upgrade logs and issue trackers from open-source repositories on GitHub \cite{github_apps}. Our approach involves systematically filtering and extracting relevant projects based on the descriptions in their change logs and issue reports, specifically focusing on entries related to GUI defects. This targeted filtering ensures that the selected applications contain genuine and relevant defects within their user interfaces.

For each identified project, we document essential application details, such as the version history, defect type, and issue severity, to build a comprehensive profile of the detected defects. To validate these defects, we carefully reproduce the reported issues, ensuring that the GUI defects are replicable and align with the descriptions provided by the developers. By leveraging this methodology, we not only ensure the relevance and authenticity of the collected data but also capture a wide variety of defect types reflective of real-world scenarios. These defects, composing of subtle layout misalignment to critical interaction failures, offer a diverse testing ground for evaluating the automated GUI Testing framework.

\subsection{Applications with injected defects}

Real-world applications exhibit a wide range of complex and unpredictable GUI defects, making it difficult to ensure consistency across testing scenarios. To address this, we inject defects at the source code level or use the MutAPK tool\cite{escobar2019mutapk, escobar2020mutapk} to introduce controlled, predefined GUI defects into mobile applications\cite{googleplay, f-droid}. The injection of specific defects allows us to maintain strict control over the testing framework's results. Introducing defects in various areas not only ensures consistency but also creates diverse fault scenarios. The controlled nature of this method enables repeatable experiments, helping researchers systematically explore the strengths and weaknesses of different testing models. Additionally, by increasing the complexity of the injected defects, we can push the boundaries of the agent, ensuring it can handle the kind of diverse challenges in real-world applications.

\subsection{Synthetic defect datasets}

For most commercial applications, source code is proprietary and public releases are generally stable, having undergone multiple testing iterations. Consequently, these apps rarely contain the early-stage GUI defects essential for benchmarking the agent. To overcome this limitation, we adopt a synthetic approach, transforming screenshots of stable applications to simulate a variety of visual and interaction defects. This technique allows us to obtain GUI defect data from any app, even complex and mature commercial applications. Specific defect construction types and examples are shown in Figure \ref{fig:synthesis_data}.

%% file: sec/4_Correlation.tex
\section{Correlation Analysis Between General Capabilities and GUI Autonomous Test Performance}

A key assumption in our approach is that a fine-tuned model performs better on specific sub-tasks because it has improved mastery of the skills needed for those tasks. In automated GUI Testing, tasks like test case generation and defect detection require both perception (\eg, recognizing visual elements) and reasoning (\eg, interpreting navigation logic and workflows). However, it's difficult to delineate exactly which capabilities contribute most to success. The ability to navigate between screens, detect overlapping elements, or respond correctly to unresponsive buttons may draw on multiple, interconnected competencies. Often, these dependencies are not straightforward.

Therefore, we propose an evaluation method that fine-tunes models on datasets specific to GUI Testing tasks and then assesses their performance on broad, standardized benchmarks. This comparative analysis,contrasting a model's pre- and post-fine-tuning performance, offers insights into the capabilities that are most relevant to specific stages of the GUI Testing process. For example, improvements in benchmarks focused on perception may indicate the model’s enhanced ability to identify subtle layout issues, while gains in reasoning-oriented benchmarks might reflect better handling of navigation errors or task flows. Additionally, real-world GUI Testing tasks often suffer from data scarcity. Our method provides a pathway for expanding datasets by strategically selecting general datasets aligned with the task’s requirements. 

This framework ties back to the core goal of our paper: building a comprehensive, end-to-end benchmark for GUI automation testing. By bridging the gap between general benchmark performance and task-specific outcomes, we offer a practical methodology for identifying the key capabilities that matter most. This not only enhances the reliability of visual-based agent for automated GUI Testing but also lays the foundation for continuous model improvement.


\begin{table*}[t]
    \vspace{-1.em}
    \small
    \tablestyle{1.8pt}{1.05} 
    \begin{tabular}{ry{40}|y{48}|y{48}y{48}y{48}|y{48}y{48}y{50}c}
        & & \multicolumn{1}{c|}{\textbf{Test Intention Generation}} & \multicolumn{3}{c|}{\textbf{Test Task Execution}} & \multicolumn{3}{c}{\textbf{GUI Defect Detection}} \\
        \multicolumn{2}{c|}{Model} & \multicolumn{1}{c|}{\textbf{Coverage}} & \multicolumn{1}{c}{\textbf{TM}} & \multicolumn{1}{c}{\textbf{EM}} & \multicolumn{1}{c|}{\textbf{SR}} & \multicolumn{1}{c}{\textbf{Accuracy}} & \multicolumn{1}{c}{\textbf{Recall-D}} & \multicolumn{1}{c}{\textbf{Recall-N}} & \\
        \shline
        \tagfont{LlaMA3-8B} & LLaVA & \ressup{17.12}{} & \ressup{31.70}{} & \ressup{4.40}{} & \ressup{3.28}{} & \ressup{24.90}{} & \ressup{6.00}{} & \ressup{69.0}{} & \\
        \tagfont{Qwen2-7B} & Qwen2-VL & \ressup{20.37}{} & \ressup{35.40}{} & \ressup{13.20}{} & \ressup{11.73}{} & \ressup{30.10}{} & \ressup{0.14}{} & \reshl{100.0}{+}{20} & \\
        \tagfont{3.5-Sonnet} & Claude & \ressup{36.18}{} & \reshl{48.88}{+}{7.3} & \reshl{21.50}{+}{2.1} & \reshl{20.45}{+}{4.3} & \ressup{7.3}{} & \ressup{10.29}{} & \ressup{0.33}{} & \\
        \tagfont{2024-02-01} & GPT-4o & \reshl{37.01}{+}{0.8} & \ressup{41.60}{} & \ressup{19.40}{} & \ressup{16.14}{} & \reshl{33.80}{+}{3.7} & \reshl{14.00}{+}{3.7} & \ressup{80.0}{} &
    \end{tabular}
    \caption{\textbf{Comparison of Different Multimodal Large Language Models as Agents for Autonomous GUI Testing.} The green numbers represent the difference between the best result and the second-best result. TM, EM, and SR denote Type Match, Exact Match, and Success Rate, respectively. Since several models tend to respond with ``no defect" during GUI defect detection, we selected both defective and non-defective data to calculate the recall metrics, denoted as Recall-D and Recall-N, respectively.}
    \label{tab:1}
    \vspace{-1.em}
\end{table*}



\begin{table}[!t]
    \centering
    \tablestyle{2pt}{1.1}
    \begin{tabular}{ry{46}|x{38}x{38}x{38}c}
        \multicolumn{2}{c|}{} & \multicolumn{3}{c}{\fontsize{7.5pt}{1em}\selectfont\textbf{Task Execution on different instructions}} & \\
        \multicolumn{2}{c|}{Model} & GPT-4o & LLaVA  & Qwen2-VL \\
        \shline
        \tagfont{} & \fontsize{7.5pt}{1em}\selectfont\textbf{Test Intention} & \ressup{16.14}{} & \ressup{3.28}{} 
 & \ressup{11.73}{} & \\      
        \tagfont{} & \fontsize{7.5pt}{1em}\selectfont\textbf{Test Steps} & \ressup{16.39}{} & \ressup{6.81}{} & \ressup{19.80}{} & 
    \end{tabular}
    \caption{\textbf{SR of VLLMs on Different Instructions for Test Task Execution.} We use two types of instructions, test intention and test steps, to compare the task completion performance of the models under each instruction type.}
    \label{tab:2}
\end{table}





%% file: sec/5_Experiment.tex
\section{Experiment}
\subsection{Experiment Setup and Evaluation Criteria}
In alignment with the workflow detailed in Section \ref{agent_workflow}, our experimental setup benchmarks the performance of various multimodal large models under a unified framework. This approach allows for a direct comparison of these models in a consistent architecture, assessing their effectiveness as agents in generating test intentions, executing test steps, and conducting GUI defect detection. Since the test case comprises a test intention and corresponding test steps, which relies on task execution result. Therefore, our primary focus is on evaluating the generation of test intentions and, based on this, assessing the effectiveness of task execution.

\textbf{Coverage.} The model generates a variable number of test intentions based on the app and background information. To account for the semantic ambiguity in test intentions, each generated intention is matched to the human-annotated ground truth by GPT as judge, assessing if it aligns with the true set of test intentions. The judgment prompt template we used is provided in Appendix. The proportion of correctly aligned intentions represents the coverage rate for test intention generation.

\textbf{TM, EM and SR.} For Test Task Execution, we employ \textbf{TM} (Type Match), \textbf{EM} (Exact Match), and \textbf{SR} (Success Rate) as evaluation metrics. For each defined tuple $(state_b, action,state_a)$, \textbf{TM} indicates whether the model correctly predicts the type of action to take in the next step. \textbf{EM} assesses, given a correct action type, whether the action details are accurate. For instance, in test step \texttt{click(3)}, where 3 is the element ID within the image, the model needs to identify the correct element to click. For a test task, if the action is entirely correct for each tuple, we consider the agent to have successfully completed the task. \textbf{SR} represents the percentage of tasks that the model successfully completes out of all tasks.

\textbf{Accuracy and Recall.} For GUI defect detection, we evaluate accuracy and recall metrics. The test set includes both data with GUI defects and normal data (represented in our defined triplet form). For \textbf{Accuracy}, we calculate the proportion of correct judgments made by the model across all data. Recall is further divided into two metrics: $Recall_{defect}$, which measures the model’s correct judgment rate on data with GUI defects, and $Recall_{no defect}$, which assesses the model's performance on normal data. These three metrics allow us to thoroughly analyze the performance of different multimodal large language models in the task of GUI defect detection.

\subsection{Baseline Autonomous GUI Testing}
To establish a baseline, we select models with significantly different architectures and native multimodal capabilities for comparison, including GPT-4o, Claude, LLaVA, and Qwen2-VL.

The experimental results, shown in Table \ref{tab:1}, indicate that GPT-4o and Claude perform comparably across most metrics and outperform open-source models. Although the difference in metrics between closed-source commercial models and open-source models is relatively small, the overall poor performance highlights a significant gap between open-source and closed-source models. In the test intention generation metric, LLaVA and Qwen2-VL lag considerably behind GPT-4o and Claude, suggesting that tasks requiring extensive knowledge and imaginative capabilities still necessitate models with sufficiently large parameters to produce richer responses. The differences in TM and EM performance further support this point: while open-source models can answer some matching tasks correctly, LLaVA struggles significantly in higher-difficulty tasks involving exact matches, whereas Qwen2-VL performs slightly better due to prior training on GUI data.

For GUI defect detection, we use a mix of data with and without defects to simulate the agent performing defect detection tasks in a real-world scenario. Since Qwen2-VL predominantly responds with ``no defect", it achieves a high recall rate on normal data but performs poorly on data with actual defects. In contrast, Claude tends to think there exists GUI defects in the data. This outcome underscores the necessity of evaluating both metrics for a comprehensive assessment.

\subsection{Ablation Study on Test Task Execution}
In Section \ref{test_case_generation}, we discussed how the action sequence performed by the agent under the guidance of a test intention can be used to construct test steps, thereby generating a complete test case. This raises the question: Does using test steps as instructions enable the agent to complete tasks more effectively compared to using the test intention? To explore this, we concatenated the action sequences within individual test tasks to form the corresponding test steps for each task. We then evaluated the performance of GPT-4o, LLaVA, and Qwen2-VL on test task execution under these two types of instructions.

The experimental results in Table \ref{tab:2} show that both LLaVA and Qwen2-VL exhibited performance improvements when given test steps as instructions. GPT-4o, however, showed minimal change in performance, even performing worse than Qwen2-VL when using test steps as instructions. This suggests that the limitation in GPT-4o's performance in test task execution does not stem from its ability to accurately interpret test intentions but rather from the inherent complexity of GUI interfaces. Qwen2-VL, having been trained on GUI data, benefits from the clarity in identifying the next action to execute, resulting in a more significant performance boost when provided with explicit test steps.


%% file: sec/6_Related_Work.tex
\section{Related Work}
\noindent\textbf{Agent on GUI navigation task.} Early GUI agents \cite{gur2018learning, shi2017world, li2020mapping, li2022spotlight} primarily relied on training models to explore GUI environments with task completion as the main objective. With advances in multimodal large models \cite{achiam2023gpt, anthropic2024introducing}, current approaches have shifted from traditional training to using techniques like prompt tuning \cite{lester2021power} and in-context learning \cite{zheng2023synapse} to guide these models in exploration tasks. AppAgent \cite{zhang2023appagent}, Mobile-Agent \cite{wang2024mobile}, and AutoDroid \cite{wen2023empowering} utilize LLMs to interpret natural language descriptions and transform them into GUI actions. Additionally, some work \cite{cheng2024seeclick, hong2024cogagent} has focused on fine-tuning large models on GUI-specific data to improve their performance in GUI environments. Agent workflow memory\cite{wang2024agent} represents another recent innovation, enhancing agents’ ability to automatically construct workflows, thus introducing a new paradigm for GUI task automation.

There has also been progress in creating benchmarks for GUI navigation \cite{shi2017world, li2020mapping, deng2024mind2web}. WebArena \cite{zhou2023webarena}, for instance, constructs realistic web environments with callable tools to study the limitations of models like GPT-4V, revealing a significant gap in agent performance compared to humans in complex tasks. AitW \cite{rawles2024androidinthewild} collected large-scale data by having annotators operate apps in a simulator to capture human instruction-following behavior, though data quality remains a concern. Building on this, AitZ \cite{zhang2024android} introduced high-quality navigation data with GPT-4-annotated Chain-of-Thought (CoT) reasoning, along with new metrics to evaluate agent performance in GUI navigation tasks.
\\
\noindent\textbf{GUI Defect Detection.} Given the close connection between GUI quality and user experience, various methods have been developed to detect bugs in GUIs. GUI Testing in industry relies heavily on scripted tests to automate function validation. To address this, AppFlow \cite{hu2018appflow} applies machine learning to identify screen components, allowing testers to develop modular libraries for core application functions. CoSer \cite{cao2024comprehensive} constructs UI state transition graphs from source code and scripts to repair outdated tests. Recently, LLMs have emerged as powerful tools in GUI Testing due to their extensive training on diverse data and strong reasoning abilities. For example, QTypist \cite{liu2023fill} focuses on generating semantic text inputs for form fields to improve exploration coverage. GPTDroid \cite{liu2024make} extracts page and widget information from the UI hierarchy, using it to create human-like interactions. AUITestAgent\cite{hu2024auitestagent} developed an industry-applicable automatic natural language-drifillven GUI Testing method. VisionDroid \cite{liu2024vision} addresses non-crash bug detection in GUIs by leveraging LLMs to detect unexpected behaviors, particularly in scenarios where testing oracles are lacking.

%% file: sec/7_Conclusion.tex
\section{Conclusion}
This paper introduces a formalized framework for Autonomous GUI Testing, aimed at addressing key limitations in visual-based agent evaluation. By structuring the testing workflow with precise mathematical definitions and decoupling GUI defect detection from task execution, we present a fair and robust environment (GTArena) for evaluating GUI Testing capabilities. Our work includes a novel data structure for capturing GUI defects, which facilitates the creation of large-scale datasets.

Furthermore, we propose a unified benchmark to assess visual-based agents equipped with multimodal large models (MLLMs), evaluating their performance across core components: test case generation, test task execution, and GUI defect detection. Through this structured benchmark, we reveal notable performance gaps between current agents and practical application for mainstream VLLMs, underscoring the need for targeted model improvements. Additionally, our methodology offers a systematic approach for fine-tuning models on task-specific datasets, while evaluating their general capabilities on broader benchmarks.

In conclusion, our work provides a fair, unified and end-to-end environment for automated GUI Testing, enabling convenient and reproducible evaluation of various multimodal large models in their role as agents. By bridging the gap between theoretical frameworks and practical evaluations, we aim to accelerate the development of more capable, reliable, and efficient agents for GUI Testing applications.

%% file: sec/Appendix.tex

\tcbset{
  myboxstyle/.style={
    colback=gray!20,     
    colframe=black!70,    
    coltitle=white,       
    fonttitle=\bfseries,  
     fontupper=\itshape,
    boxrule=0.8mm,       
    arc=1mm,               
    boxsep=1mm,             
    left=1mm,              
    right=1mm,              
    top=1mm,               
    bottom=1mm,             
    toptitle=0mm,           
    bottomtitle=0mm,     
    enhanced,                  
  }
}

\section{GUI Defect Types}\label{sec:GUI_defect_types}
The specific types of GUI defects and examples are shown in Table \ref{table: Display Defects} and \ref{table: Interaction Defects}.
\begin{table*}[!htbp]
\centering
\begin{tabular}{p{1.7cm}|p{2cm}|p{7.5cm}|p{5.5cm}}
\hline
\textbf{Defect  Categories} & \textbf{Defect} & \textbf{Description} & \textbf{Example} \\
\hline
\multirow{2}{*}{Display} 
& Content Error & This defect involves text that appears as garbled or unintelligible characters on the screen, making information difficult to read or understand. & Replace content in string.xml with `null'. \\
\cline{2-4}
& Data Type or Format Error & This defect occurs when data is displayed in inappropriate or unexpected formats, which can lead to misinterpretation or difficulty in understanding the data. & Letters are allowed to be entered in the date input field. The page shows the date ``2021-06-15" as ``20210615". \\
\hline
\multirow{4}{*}{Layout} 
& UI Element Missing & This defect refers to the absence of crucial UI elements within the interface, which can hinder user interaction or functionality. & Image not loaded or displayed broken. ``New" page lacks a save button. \\
\cline{2-4}
& UI Element Overlapping & This defect describes scenarios where UI components overlap one another, obscuring content and potentially making certain functions inaccessible. & The labels for ``Total Expenditure" and ``Remaining Budget" overlap. \\
\cline{2-4}
& Alignment Issue & This defect is identified when UI elements are not properly aligned, leading to a visually disorganized interface that can detract from user experience. & In a center-aligned navigation bar, one item is right-aligned. \\
\cline{2-4}
& Uneven Spacing & This defect is characterized by irregular spacing between UI elements, which can create a cluttered or unbalanced appearance, affecting the aesthetic and usability. & Two elements are spaced too far apart, resulting in a large area of whitespace. \\
\hline
\multirow{3}{*}{Style} 
& Inconsistent Color & This defect arises when the color scheme of UI elements is mismatched or poorly chosen, potentially leading to a visually unappealing or confusing interface. & Most of the icon colors in the navigation bar are the same, with a few exceptions. \\
\cline{2-4}
& Inconsistent Element Size & This defect pertains to UI elements that vary significantly in size, which can confuse users and disrupt the visual flow of the application, affecting usability. & Some fonts are too large while others are too small. \\
\cline{2-4}
& Abnormal UI Element State & This defect involves UI elements that display unexpected behaviors or appearances when they are interacted with, such as being clicked or focused, which can confuse users or hinder interaction. & The submit button appears in an active state although it is not being clicked. \\
\hline
\end{tabular}
\caption{UI Display Defects}
\label{table: Display Defects}
\end{table*}

\begin{table*}[ht]
\centering
\begin{tabular}{p{1.7cm}|p{2cm}|p{7.5cm}|p{5.5cm}}
\hline
\textbf{Defect Categories} & \textbf{Defect} & \textbf{Description} & \textbf{Example} \\
\hline
\multirow{2}{*}{Operation} 
& Operation No Response & This defect occurs when there is no feedback or action following user interactions, leading to uncertainty and frustration for the user. & Clicked submit button but there was no response. There are more content below, but could not scroll down when the user swipes down. \\
\cline{2-4}
& Virtual Keyboard Related Issue & This defect involves problems with the virtual keyboard that affect typing or input, such as unexpected behavior or layout issues. & The virtual keyboard cannot wake up automatically. \\
\hline
\multirow{2}{*}{Task} 
& Navigation Logic Error & This defect refers to flaws in the navigation logic that result in incorrect or unintended application flows, potentially leading users to incorrect destinations or functions. & Click the 'Default Setting' but jump to 'UI interface'. \\
\cline{2-4}
& Unexpected Task Result & This defect occurs when the results of tasks do not align with the anticipated results or specifications, leading to confusion and potential errors in usage. & Theme change not working. The recipe could not be deleted. \\
\hline
\end{tabular}
\caption{UI Interaction Defects}
\label{table: Interaction Defects}
\end{table*}

\section{GUI Defect Dataset Examples}\label{sec:GUI Defect Dataset}

\begin{figure*}[!t]
    \centering
    \includegraphics[width=16cm]{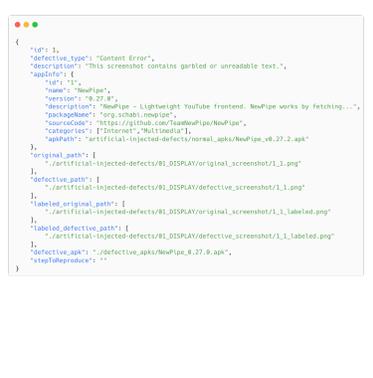}
    \caption{Example Defects in Artificial Injected Data.}
    \label{fig:example_artificial_injected}
\end{figure*}

\begin{figure*}[!t]
    \centering
    \includegraphics[width=16cm]{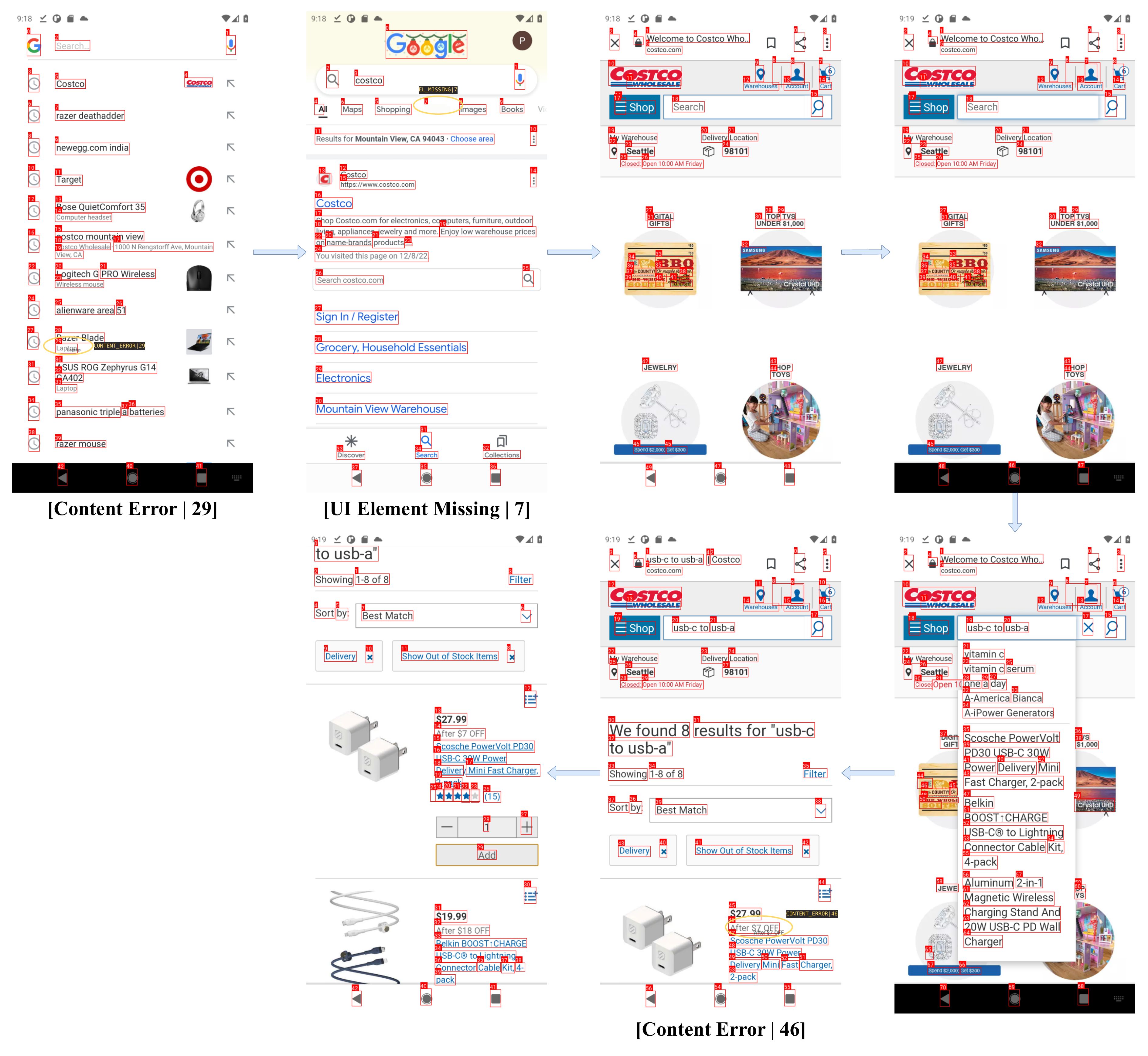}
    \caption{Example episode form the AitW\_with\_Defects.}
    \label{fig:example_aitw}
\end{figure*}

Some real-world defects from Github releases in Table \ref{table:real_world_defects}. Examples of artificial injected defects and episode from AitW with defects are show in Figure \ref{fig:example_artificial_injected} and Figure \ref{fig:example_aitw}.
\begin{table*}[!htbp]
\centering
\begin{tabular}{p{1.2cm}|p{7.6cm}|p{7.6cm}}
\hline
\textbf{Release} & \textbf{Display Defect} & \textbf{Interaction Defect} \\
\hline
\href{https://github.com/krille-chan/fluffychat/releases/tag/v1.21.0}{v1.21.0} 
& - Stickers from Gboard have black background (fixed) \newline
- mxc reactions not rendered correctly (fixed) 
& - Broken localization with empty strings in it (fixed) \\
\hline
\href{https://github.com/krille-chan/fluffychat/releases/tag/v1.17.2}{v1.17.2} 
& / 
& - Add cancel button to key request dialog \newline
- Encode component for links correctly \newline
- Forward arbitrary message content \newline
- Open public room bottom sheet by alias \\
\hline
\href{https://github.com/you-apps/VibeYou/releases/tag/v3.0}{v3.0} 
& - Song placeholder icon in player view 
& / \\
\hline
\href{https://github.com/you-apps/VibeYou/releases/tag/v2.0}{v2.0} 
& - Launcher icon background color 
& - Disable favourite button for local songs \\
\hline
\href{https://github.com/you-apps/VibeYou/releases/tag/v1.0}{v1.0} 
& - Color of status and navigation bar \newline
- Splash screen background color in dark mode 
& / \\
\hline
\href{https://github.com/RetroMusicPlayer/RetroMusicPlayer/releases/tag/v6.0.0}{v6.0.0} 
& / 
& - Top/Recent Artists/Albums not updating (Wrong sort order) \newline
- All Blacklist related crashes \newline
- Restart button not working in crash activity \\
\hline
\href{https://github.com/RetroMusicPlayer/RetroMusicPlayer/releases/tag/v5.8.4}{v5.8.4} 
& / 
& - Crash when adding folders to blacklist \\
\hline
\href{https://github.com/RetroMusicPlayer/RetroMusicPlayer/releases/tag/v5.8.3}{v5.8.3} 
& - Incorrect song data in notification 
& / \\
\hline
\href{https://github.com/RetroMusicPlayer/RetroMusicPlayer/releases/tag/v5.8.0}{v5.8.0} 
& / 
& - Settings change not reflecting immediately \newline
- Crash when clicking on Playlist in the Search Tab \\
\hline
\href{https://github.com/RetroMusicPlayer/RetroMusicPlayer/releases/tag/v5.6.0}{v5.6.0} 
& - Incorrect colors when no cover art is available \newline
- Blank album cover bug 
& - Lockscreen dragging glitch \newline
- Favorite not updating when song is changed \newline
- Playlist not getting created \& playlist creation crash with same name \newline
- Bug in ``Plain” Now playing theme where onClick event is consumed by the views behind the bottom sheet \\
\hline
\end{tabular}
\caption{Example of real-world defects description}
\label{table:real_world_defects}
\end{table*}

